\pdfoutput=1

\documentclass[11pt]{article}

\usepackage[]{acl}

\usepackage{times}
\usepackage{latexsym}
\usepackage{graphicx}
\usepackage{booktabs}
\usepackage{microtype}
\usepackage{tikz-dependency}
\usepackage{booktabs}
\usepackage{multirow}
\usepackage{xspace}
\usepackage{xcolor}
\usepackage{hyperref}
\usepackage{subcaption}
\usepackage{relsize}
\usepackage{amsmath}
\usepackage{comment}
\usepackage{listings}

\newcommand\F{{$\text{F}_1$}\xspace}

\newcommand{\norec}{\textbf{NoReC}\xspace}
\newcommand{\norecfine}{\textbf{NoReC$_{fine}$}\xspace}

\newcommand{\cat}{\textbf{CA}\xspace}
\newcommand{\basque}{\textbf{EU}\xspace}
\newcommand{\dsu}{\textbf{DSU}\xspace}
\newcommand{\mpqa}{\textbf{MPQA}\xspace}
\newcommand{\todo}[1]{\colorbox{red}{#1}}

\newenvironment{cenumerate}{\begin{list}{\labelenumi}{\usecounter{enumi}\topsep=.2\smallskipamount\itemsep=0pt\parsep=1pt\labelwidth=1.0em}}{\end{list}}

\makeatletter
\newcommand\crulefill{\leavevmode
\begingroup 
\setlength{\dimen@}{0.5ex}
\addtolength{\dimen@}{0.4pt}
\leaders\hrule height\dimen@ depth -.5ex \hfill
\endgroup
\kern\z@}

\usepackage[T1]{fontenc}

\usepackage[utf8]{inputenc}

\usepackage{microtype}

\title{Direct parsing to sentiment graphs}
\author{David Samuel,$^1$ Jeremy Barnes,$^2$ Robin Kurtz,$^3$ Stephan Oepen,$^1$ \\
  \textbf{Lilja Øvrelid$^1$ and Erik Velldal$^1$}\vspace{0.25em} \\
  $^1$University of Oslo, Language Technology Group \\
  $^2$University of the Basque Country UPV/EHU, HiTZ Center -- Ixa \\
  $^3$National Library of Sweden, KBLab\vspace{0.25em} \\
  \texttt{\{davisamu, oe, liljao, erikve\}@ifi.uio.no} \\
  \texttt{jeremy.barnes@ehu.eus, robin.kurtz@kb.se} \\}

\begin{document}
\maketitle
\begin{abstract}
This paper demonstrates how a graph-based semantic parser can be applied to the task of structured sentiment analysis, directly predicting sentiment graphs from text. We advance the state of the art on 4 out of 5 standard benchmark sets. 
We release the source code, models and predictions.\footnote{\url{github.com/jerbarnes/direct_parsing_to_sent_graph}}
\end{abstract}

\section{Introduction}
\begin{figure}[]
\centering
\includegraphics[width=\columnwidth]{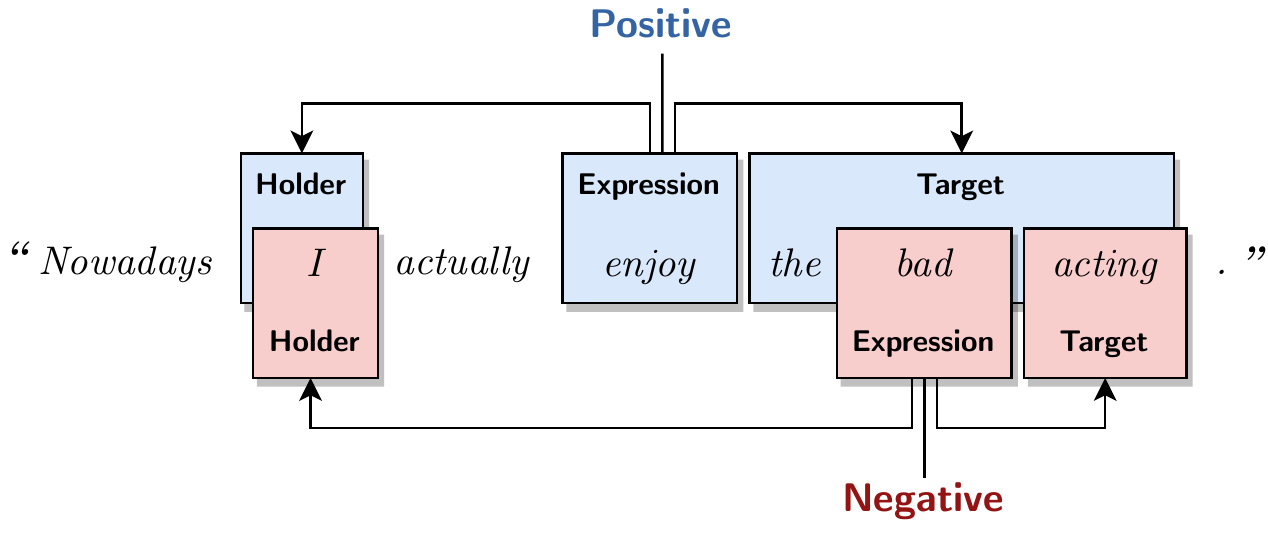}
\caption{A sentiment graph for the phrase \textit{``Nowadays I actually enjoy the bad acting,''} which contains an example of nesting of two opposing opinions.}
\label{fig:ssa}
\end{figure}

The task of structured sentiment analysis (SSA) is aimed at locating all \textit{opinion tuples} within a sentence, where a single opinion contains a) a polar expression, b) an optional holder, c) an optional sentiment target, and d) a positive, negative or neutral polarity, see Figure~\ref{fig:ssa}. 
While there have been sentiment corpora annotated with this information for decades \cite{Wiebe2005,toprak-etal-2010-sentence}, there have so far been few attempts at modeling the full representation, rather focusing on various subcomponents, such as the polar expressions and targets without explicitly expressing their relations \cite{peng2019knowing,xu-etal-2020-position} or the polarity \cite{yang-cardie-2013-joint, katiyar-cardie-2016-investigating}.  

Dependency parsing approaches have recently shown promising results for SSA \cite{barnes-etal-2021-structured, peng2021sparse}. Here we present a novel sentiment parser which, unlike previous attempts, predicts sentiment graphs directly from text without reliance on heuristic lossy conversions to intermediate dependency representations. 
The model takes inspiration from successful work in meaning representation parsing, and in particular the permutation-invariant graph-based parser of
\newcite{samuel-straka-2020-ufal} called PERIN.

Experimenting with several different graph encodings, we evaluate our approach on five datasets from four different languages, and find that it compares favorably to dependency-based models across all datasets; most significantly on the more structurally complex ones -- \norec and \mpqa.

\section{Related work}

 Proposing a dependency parsing approach to the full task of SSA, \newcite{barnes-etal-2021-structured} 
 show that it leads to strong improvements over state-of-the-art baselines.
\newcite{peng2021sparse} propose a sparse fuzzy attention mechanism to deal with the sparseness of dependency arcs in the models from \newcite{barnes-etal-2021-structured} and show further improvements. 
However, in order to apply the parsing algorithm of \citet{dozat-manning-2018-simpler}, both of these approaches have to rely on a \textit{lossy} conversion to bi-lexical dependencies with ad-hoc internal head choices for the nodes of the abstract sentiment graph, see Section \ref{sec:dependency-problems} for a discussion of these issues.


More generally, decoding structured graph information from text has sparked a lot of interest in recent years, especially for parsing meaning representation graphs \citep{oepen-etal-2020-mrp}. There has been tremendous progress in developing complex transition-based and graph-based parsers \cite{hershcovich-etal-2017-transition, mcdonald-pereira-2006-online, dozat-manning-2018-simpler}. In this paper, we adopt PERIN \cite{samuel-straka-2020-ufal}, a state-of-the-art graph-based parser capable of modeling a superset of graph features needed for our task.


\section{Issues with dependency encoding}
\label{sec:dependency-problems}

As mentioned above, previous dependency parsing approaches to SSA have relied on a \textit{lossy} bi-lexical conversion. This is caused by an inherent ambiguity in the dependency encoding of two nested text spans with the same head (defined as either the first or the last token in \newcite{barnes-etal-2021-structured}).

\begin{figure}[t]
    \centering
    \includegraphics[width=\columnwidth]{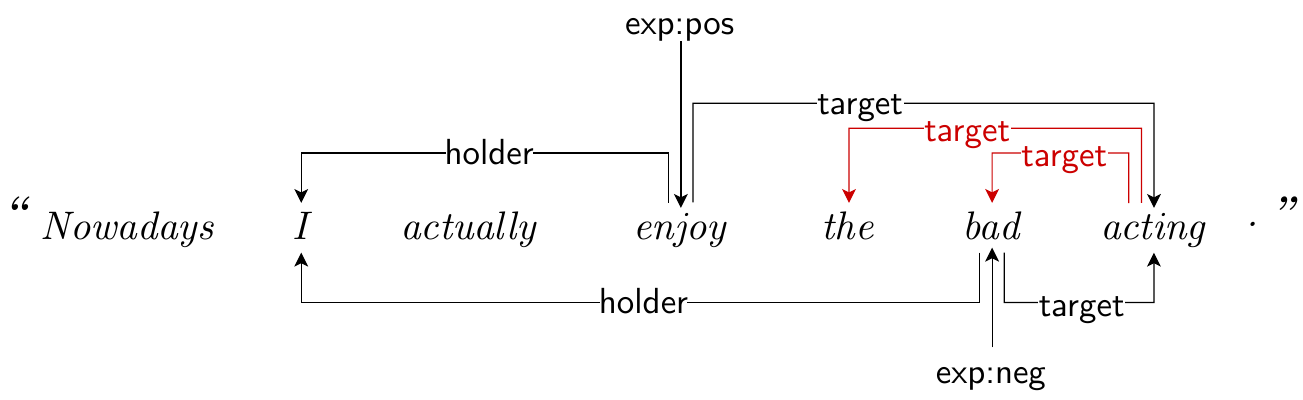}
    \caption{Ambiguous \textit{targets} when encoding the sentence \textit{``Nowadays I actually enjoy the bad acting''} as a head-final bi-lexical dependency graph \citep{barnes-etal-2021-structured}.}
    \label{fig:dependency-error}
\end{figure}

To be concrete, we can use the running example \textit{``Nowadays I actually enjoy the bad acting,''} which has two opinions with nested targets; \textit{``the bad acting,''}, which is associated with a positive polarity indicated by the polar expression \textit{``enjoy''}, and \textit{``acting,''}, with a negative polarity expressed by \textit{``bad''}. As shown in the dependency representation in Figure~\ref{fig:dependency-error}, both expression--target edges correctly lead to the word \textit{``acting''} but it is impossible to disambiguate the prefix of both targets in the bi-lexical encoding, i.e., to determine that the tokens  \textit{``the''} and \textit{``bad''} are part of the target only for the positive opinion. For that, we need a more abstract graph encoding, such as the ones suggested in this paper.

\section{PERIN model}

PERIN is a general permutation-invariant text-to-graph parser. The output of the parser can be a directed graph with labeled nodes connected by labeled edges where each node is anchored to a span of text (possibly empty or discontinuous). We propose three graph representations for SSA that meet these constrains and thus can be easily modeled by this parser.

We use only a subset of the full PERIN's functionality for our SSA version -- it does not need to use the ``relative label rules'' and model node properties or edge attributes. Please consult the original work for more technical details about PERIN \cite{samuel-straka-2020-ufal}.

\subsection{Architecture}

\begin{figure}[t]
    \centering
    \includegraphics[width=\columnwidth]{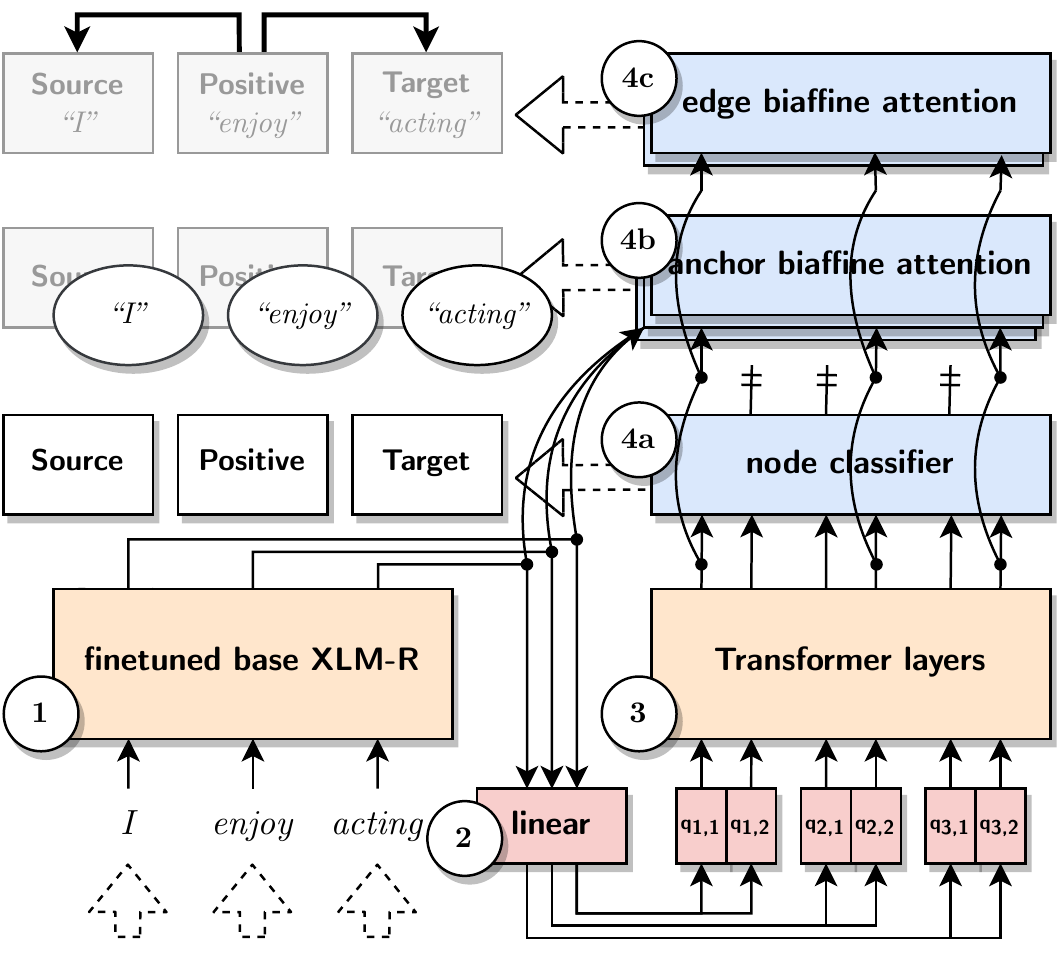}
    \caption{Diagram of the PERIN architecture; 1)~each token gets a contextualized embedding and 2)~generates queries, 3)~queries are further processed and 4)~they are put through a)~node, b)~anchor and c)~edge classification heads. 
    }
    \label{fig:architecture}
\end{figure}

PERIN processes the input text end-to-end in four steps, illustrated in Figure~\ref{fig:architecture}: 1)~To encode the input, PERIN uses contextualized embeddings from XLM-R \cite[base size; ][]{conneau-etal-2020-unsupervised} and combines them with learned character-level embeddings;\footnote{The character embeddings are not discussed in the PERIN description paper but they are included in the official implementation. We use a single bidirectional GRU layer to process the characters of each token and add the result to the contextualized embeddings. Note that we also excluded them from Figure \ref{fig:architecture} to simplify the illustration.} 2)~each token is mapped onto latent \textit{queries} by a linear transformation; 3)~a stack of Transformer (encoder) layers without positional embedding \cite{NIPS2017_3f5ee243} optionally models the inter-query dependencies; and 4)~classification heads select and label queries onto nodes, establish anchoring from nodes to tokens, and predict the node-to-node edges.


\subsection{Permutation-invariant query-to-node matching}

Traditional graph-based parsers are trained as autoregressive sequence-to-sequence models. PERIN does not assume any prior ordering of the graph nodes.\footnote{Permutation invariance is arguably more important for semantic graphs (with abstract nodes) than for the sentiment graphs. Yet, in case of nested nodes, there is no apparent order, so we do not constrain the model by any ordering assumptions.} Instead, it processes all queries in parallel and then dynamically maps them to gold nodes.

Based on the predicted probabilities of labels and anchors, we create a weighted bipartite graph between all queries and nodes. The goal is to find the most probable matching, which can be done efficiently in polynomial time by using the Hungarian algorithm. Finally, every node is assigned to a query and we can backpropagate through standard cross-entropy losses to update the model weights.

\begin{figure*}[ht]
    \centering
    \includegraphics[width=\textwidth]{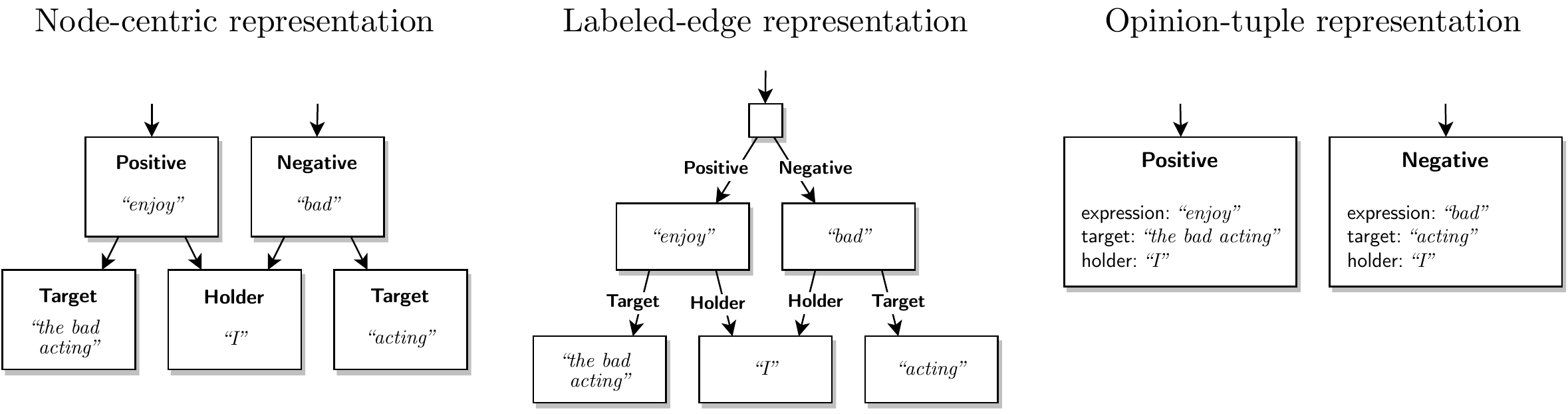}
    \caption{Three representations of the structured sentiment graph for sentence \textit{``Nowadays I actually enjoy the bad acting.''}}
    \label{fig:graph-representations}
\end{figure*}

\subsection{Graph encodings}

PERIN defines an overall framework for general graph parsing, it can cater to specific graph encodings by changing the subset of its classification heads. 
In parsing the abstract sentiment structures, there are several possible lossless graph encodings depending on the positioning of the polarity information and the sentiment node type. We experiment with three variations (Figure~\ref{fig:graph-representations}) and later show that while the graph encoding improves performance, this improvement largely depends on the type of encoding used.
\begin{cenumerate}
    \item \textbf{Node-centric encoding}, with labeled nodes and directed unlabeled arcs. Each node corresponds to a target, holder or sentiment expression; edges form their relationships. The parser uses a multi-class node head, an anchor head and a binary edge classification head.
    \item \textbf{Labeled-edge encoding}, with deduplicated unlabeled nodes and labeled arcs. Each node corresponds to a unique text span from some sentiment graph, while edge labels denote their relationships and functions. The model has a binary node classifier, an anchor classifier and a binary and multi-class edge head.
    \item \textbf{Opinion-tuple encoding}, which represents the structured sentiment information as a sequence of opinion four-tuples. This encoding is the most restrictive, having the lowest degrees of freedom. The parser utilizes a multi-class node head and three anchor classifiers, it does not need an edge classifier.
\end{cenumerate}



\section{Data}
Following \citet{barnes-etal-2021-structured} we employ five structured sentiment datasets in four languages, the statistics of which are shown in Table~\ref{tab:statistics}.
The largest dataset is the \norecfine dataset \cite{ovrelid-etal-2020-fine}, a multi-domain dataset of professional reviews in Norwegian.  \basque and \cat  \cite{barnes-etal-2018-multibooked} contain  hotel reviews in Basque and Catalan, respectively. \mpqa \cite{Wiebe2005} annotates news wire text in English. Finally, \dsu \cite{toprak-etal-2010-sentence} annotates English reviews of online universities.  We use the SemEval 2022 releases of \mpqa and \dsu \cite{barnes-etal-2022-semeval}.\footnote{Available from \url{https://competitions.codalab.org/competitions/33556}.} 

\begin{table}[t]
    \centering
    \catcode`! = 13\def\!{\bfseries}
    \resizebox{\columnwidth}{!}{%
    \begin{tabular}{@{}llrrrrrrr@{}}
    \toprule
       & & \hspace{-1em}\!sentences & \hspace{-0.5em}\!holders & \hspace{-0.5em}\!targets & \!exps. & \!$\mathbf{+}$ & \!neu & \!$\mathbf{-}$\\
   \midrule
   \multirow{3}{*}{\norec} & train & 8634  & 898 & 6778 & 8448  & 5684& \crulefill & \hspace{-0.5em}2756 \\
   & dev & 1531 &  120 &  1152 & 1432 & 988 & \crulefill & 443\\
   & test & 1272  & 110 &  993 & 1235 & 875 & \crulefill & 358\\
    \cmidrule(l){2-9}
    \multirow{3}{*}{\cat}  & train & 1174 & 169  & 1695 & 1981   & 1272 & \crulefill &708 \\
                 & dev & 168 & 15  & 211  & 258 & 151 & \crulefill &107 \\
                 & test & 336 &  52 & 430 & 518 & 313 & \crulefill &204\\
    \cmidrule(l){2-9}
   \multirow{3}{*}{\basque} & train & 1064  & 205 & 1285 & 1684 & 1406 & \crulefill & 278\\
                  & dev & 152 & 33 & 153 & 204 & 168 & \crulefill & 36 \\
                  & test & 305 & 58  & 337 & 440 &  375 & \crulefill & 65\\
    \cmidrule(l){2-9}
  
    \multirow{3}{*}{\mpqa}   & train & 5873 & 1431 & 1487 &  1715 & 671 & 337 & 698 \\
                        & dev & 2063 & 414 &  503 & 581 & 223 & 126 & 216\\
                       & test & 2112 &  434 &  462 &518 & 159 & 82 & 223\\
    \cmidrule(l){2-9}
       \multirow{3}{*}{\dsu} &  train & 2253 & 65 &  836 & 836 & 349 & 104 & 383 \\
                    & dev & 232  & 9 & 104 & 104 & 31 & 16 & 57\\
                    & test & 318 & 12 & 142 & 142 &  59 & 12 & 71\\
                 \bottomrule
    \end{tabular}
}%
    
    \caption{Statistics of the datasets, including number of sentences per split, as well as number of holder, target, and polar expression annotations. Additionally, we include the distribution of polarity -- restricted to positive, neutral, and negative -- in each dataset.}
    \label{tab:statistics}
\end{table}

\subsection{Nested dependencies}
Returning to the issue of dependency encoding for nested elements discussed in Section \ref{sec:dependency-problems}, Table \ref{tab:nested_analysis} shows that the amount of nesting in the SSA datasets is not negligible, further motivating our abstract graph encodings for this task.

\begin{table}[t]
    \centering
    \smaller
    \begin{tabular}{@{}lrrrrrr@{}}
    \toprule
        \multirow{2}{*}{\textbf{}} & \multicolumn{2}{c}{\textbf{holders}} & \multicolumn{2}{c}{\textbf{targets}} & \multicolumn{2}{c@{}}{\textbf{expressions}} \\
        & \# & \% & \# & \% & \# & \% \\
        \midrule
        \norec &  95 & 1.5 & 1187 & 14.1 & 1075 & 9.3\\
        \basque & 30 & 2.2 & 79 & 4.5 & 16 & 0.7 \\
        \cat &  43 & 2.9 & 28 & 1.2 & 23 & 0.9\\
        \mpqa & 48 & 2.2 & 250 & 9.3 & 145 & 5.6\\
        \dsu & 0 & 0.0 & 10 & 1.1 & 7 & 0.5\\
        \bottomrule
    \end{tabular}
    \caption{Count and percentage of nesting for each dataset.}
    \label{tab:nested_analysis}
\end{table}

Table \ref{tab:arcloss}a further shows the amount of dependency edges lost because of overlap. Finally, Table \ref{tab:arcloss}b shows the S\F score when converting the gold sentiment graphs to bi-lexical dependency graphs and back -- an inherent upper bound for any dependency parser.
\begin{table}[!t]
\centering
\small
\begin{tabular}{lr}
\toprule
 \norec & 8.8\% \\
 \basque         & 4.5\% \\
 \cat         & 6.7\% \\
 \mpqa       & 4.2\% \\
 \dsu    & 0.5\% \\
\bottomrule
\end{tabular}
\hspace{1cm}
    \begin{tabular}{lllllll}
    \toprule
        \norec & 93.6 \\
        \basque & 95.2 \\
        \cat & 97.6 \\
        \mpqa & 96.6\\
        \dsu & 99.8 \\
        \bottomrule
    \end{tabular}
    \caption{a) Percentages of dependency arcs lost due to overlap; b) Sentiment Graph F1 after converting test sets to head-final and then reconverting to json format.}
\label{tab:arcloss}
\end{table}

\section{Experiments}


\subsection{Evaluation}

Following \newcite{barnes-etal-2021-structured}, we evaluate our models using Sentiment Graph \F (\textbf{S\F}). This metric considers that each sentiment graph is a tuple of  (holder, target, expression, polarity). A true positive is defined as an exact match at graph-level, weighting the overlap in predicted and gold spans for each element, averaged across all three spans. For precision it weights the number of correctly predicted tokens divided by the total
number of predicted tokens (for recall, it divides instead by the number of gold tokens). S\F allows for empty holders and targets. 

In order to further analyze the models, we also include token-level \F for extraction of Holders, Targets, and Polar Expressions, as well as Non-polar Sentiment Graph \F (\textbf{NS\F}).

\subsection{Models}

We compare our models to the head-final dependency graph parsers from \newcite{barnes-etal-2021-structured} as well as the second-order Sparse Fuzzy Attention parser of \newcite{peng2021sparse}.
For all models, we perform 5 runs with 5 different random seeds and report the mean and standard deviation. Results on development splits are provided in Appendix \ref{sec:dev-results}, training details are in Appendix \ref{sec:training-details}.

\subsection{Results}

Table \ref{tab:main_results} shows the main results. Our models outperform both dependency graph models on S\F, although the results are mixed for span extraction. The opinion-tuple encoding gives the best performance on S\F (an average of 6.2 percentage points (pp.) better than \newcite{peng2021sparse}), followed by the labeled edge encoding (3.0) and finally the node-centric encoding (2.1). 

For extracting spans, the opinion tuple encoding also achieves the best results on \norec, either labeled-edge or node centric on \cat and \mpqa,  while \newcite{peng2021sparse} is best on \basque and \dsu. This suggests that the main benefit of PERIN is at the structural level, rather than local extraction.

\renewcommand{\arraystretch}{0.9}

\begin{table*}[t]
    \newcommand{\impr}[1]{\underline{\textbf{#1}}}
    \centering
    \catcode`! = 13\def\!{\bfseries}
    \small
    \resizebox{.8\textwidth}{!}{
    \begin{tabular}{@{}llr@{}lr@{}lr@{}lr@{}lr@{}l@{}}
    \toprule
  \multirow{2}{*}{\textbf{Dataset}} & \multirow{2}{*}{\textbf{Model}} & \multicolumn{6}{c}{\crulefill~~\textbf{Span \F}~~\crulefill}   & \multicolumn{4}{c@{}}{\crulefill~~\textbf{Sent. graph}~~\crulefill} \\

    & & 
    \multicolumn{2}{l}{\text{Holder}} & \multicolumn{2}{l}{\text{Target}} & \multicolumn{2}{l}{\text{Exp.}} & \multicolumn{2}{l}{\text{NS\F $\uparrow$}} & \multicolumn{2}{l}{\text{S\F $\uparrow$}} \\
    \midrule
      
      \multirow{5}{*}{\norec}
    & \newcite{barnes-etal-2021-structured}   & 60.4 & & 54.8  & & 55.5  & & 39.2  & & 31.2\\
    & \newcite{peng2021sparse} & 63.6 & & 55.3 & & 56.1 & & 40.4 & & 31.9  \\
    & PERIN -- node-centric & 60.3 & $^{\pm1.8}$ & 51.8 & $^{\pm2.5}$ & 54.2 & $^{\pm0.9}$ & 42.7 & $^{\pm0.6}$ & 39.3 & $^{\pm0.7}$\\
    & PERIN -- labeled edge  & 64.0 & $^{\pm1.5}$ & 52.3 & $^{\pm4.2}$ & 56.1 & $^{\pm2.7}$ & 43.7 & $^{\pm2.2}$ & 40.4& $^{\pm2.1}$\\
    & PERIN -- opinion-tuple & 65.1 & $^{\pm2.5}$ & \hspace{-0.5em}*58.3 & $^{\pm1.5}$ & \hspace{-0.5em}*60.7 & $^{\pm1.1}$ & 47.8 & $^{\pm1.2}$ & \!41.6 & $^{\pm0.7}$ \\

    \cmidrule(l){2-12}
      \multirow{5}{*}{\basque}
      & \newcite{barnes-etal-2021-structured}  & 60.5  && 64.0  && 72.1   && 58.0  && 54.7  \\
      & \newcite{peng2021sparse} & 65.8 && 71.0 &&  76.7 && 66.1 && \!62.7 \\
      & PERIN -- node-centric & 58.9 & $^{\pm1.1}$ & 63.5 & $^{\pm1.5}$ & 73.9 & $^{\pm0.6}$ & 59.8 & $^{\pm0.7}$ & 58.6 & $^{\pm0.7}$\\
      & PERIN -- labeled edge  & 57.6 & $^{\pm2.5}$ & 64.9 & $^{\pm0.8}$ & 72.5 & $^{\pm1.9}$ & 60.0 & $^{\pm1.4}$ & 58.8& $^{\pm1.3}$\\
      & PERIN -- opinion-tuple & 64.2 & $^{\pm2.5}$ & 67.4 & $^{\pm0.8}$ & 73.2 & $^{\pm1.2}$ & 62.5 & $^{\pm1.2}$ & 61.3 & $^{\pm1.0}$ \\ 

     \cmidrule(l){2-12}
      \multirow{5}{*}{\cat}
      & \newcite{barnes-etal-2021-structured}  & 37.1  && 71.2  && 67.1  && 59.7  && 53.7  \\
      & \newcite{peng2021sparse} & 46.2 && 74.2 &&  71.0 && 64.5 && 59.3\\
      & PERIN -- node-centric & 56.1 & $^{\pm3.0}$ & 69.8 & $^{\pm0.4}$ & 70.5 & $^{\pm0.5}$ & 63.5 & $^{\pm0.6}$ & 61.7 & $^{\pm0.6}$\\
      & PERIN -- labeled edge  & 60.8 & $^{\pm5.1}$ & 70.8 & $^{\pm1.9}$ & 72.5& $^{\pm0.8}$ & 64.5& $^{\pm1.4}$ & 62.1& $^{\pm1.3}$\\
      & PERIN -- opinion-tuple & 48.0 & $^{\pm3.9}$ & 72.5 & $^{\pm0.7}$ & 68.9 & $^{\pm0.2}$ & 65.7 & $^{\pm0.7}$ & \!63.3 & $^{\pm0.6}$ \\

      \cmidrule(l){2-12}
      \multirow{5}{*}{\mpqa}
      & \newcite{barnes-etal-2021-structured}  & 46.3 && 49.5  && 46.0 && 26.1 && 18.8 \\
      & \newcite{peng2021sparse} & 47.9 && 50.7 && 47.8 && 38.6 && 19.1\\
      & PERIN -- node-centric & 58.4 & $^{\pm2.3}$ & 60.3 & $^{\pm2.0}$ & 55.8 & $^{\pm1.5}$ & 38.7 & $^{\pm1.6}$ & 28.3 & $^{\pm0.9}$\\
      & PERIN -- labeled edge & 53.6 & $^{\pm1.2}$ & 53.4& $^{\pm1.9}$ & 53.4& $^{\pm1.1}$ & 33.8& $^{\pm1.5}$ & 27.0& $^{\pm0.9}$\\
      & PERIN -- opinion-tuple & 55.7 & $^{\pm1.7}$ & \hspace{-0.5em}*64.0 & $^{\pm0.6}$ & 53.5 & $^{\pm1.2}$ & \hspace{-0.5em}*45.1 & $^{\pm1.1}$ & \hspace{-0.5em}*\!34.1 & $^{\pm1.1}$ \\

      \cmidrule(l){2-12}
      \multirow{5}{*}{\dsu}
    & \newcite{barnes-etal-2021-structured} & 37.4 && 42.1 && 45.5  && 34.3 && 26.5  \\
    & \newcite{peng2021sparse} & 50.0 && 44.8 && 43.7 && 35.0 && 27.4\\
    & PERIN -- node-centric & 31.4 & $^{\pm5.6}$ & 35.0 & $^{\pm1.6}$ & 35.1 & $^{\pm2.2}$ & 24.8 & $^{\pm0.7}$ & 22.9 & $^{\pm1.5}$\\
    & PERIN -- labeled edge & 32.5 & $^{\pm6.8}$ & 38.0& $^{\pm3.7}$ & 36.2& $^{\pm2.5}$ & 28.8& $^{\pm2.0}$ & 27.3& $^{\pm1.5}$\\
    & PERIN -- opinion-tuple & 42.2 & $^{\pm4.6}$ & 40.6 & $^{\pm2.7}$ & 39.3 & $^{\pm2.5}$ & 33.2 & $^{\pm2.4}$ & \!31.2 & $^{\pm2.4}$ \\
    \bottomrule
    \end{tabular}
    }
    \caption{Experiments comparing the PERIN model with previous results. We show the average values and their standard deviations from 5 runs. {\!Bold} numbers indicate the best result for the main S\F metric in each dataset.
    *~marks significant difference between our two best approaches, determined by bootstrap testing (see Appendix~\ref{sec:appendix:bootstrap}).
    }
    \label{tab:main_results}
\end{table*}

\renewcommand{\arraystretch}{1.0}

\begin{table}[tb]
    \newcommand{\impr}[1]{\underline{\textbf{#1}}}
    \centering
    \catcode`! = 13\def\!{\bfseries}
    \resizebox{\columnwidth}{!}{
    \begin{tabular}{@{}llllllr@{}l@{}}
    \toprule
    \multirow{2}{*}{\textbf{Dataset}} & \multirow{2}{*}{\textbf{Model}} & \multicolumn{3}{c}{\crulefill~~\textbf{Span \F}~~\crulefill}   & \multicolumn{3}{c@{}}{\crulefill~~\textbf{Sent. graph}~~\crulefill} \\
        & & H. & T. & E. &  NS\F\hspace{-2em} & \multicolumn{2}{l@{}}{S\F $\uparrow$} \\
    \midrule

      
      \multirow{2}{*}{\norec}
    & XLM-R dependency & 58.5 & 49.9 & 58.5 & 37.4 & 31.9 & \\
    & Frozen PERIN & 48.3 & 51.9 & 57.9 & \hspace{-0.5em}*41.8 & \hspace{-0.5em}*\!35.7 & $^{\pm0.6}$ \\

    \cmidrule(l){2-8}
      \multirow{2}{*}{\basque}
      & XLM-R dependency & 50.0 & 60.3 & 70.0 & 55.1 & 51.0 & \\
      & Frozen PERIN & 55.5 & 58.5 & 68.8 & 53.1 & \!51.3 & $^{\pm1.2}$\\

     \cmidrule(l){2-8}
      \multirow{2}{*}{\cat}
      & XLM-R dependency & 24.9 &  67.7 & 67.3 & 54.8 & 50.5 & \\
      & Frozen PERIN & \hspace{-0.5em}*39.8 & 69.2 & 66.3 & \hspace{-0.5em}*60.2 & \hspace{-0.5em}*\!57.6 & $^{\pm1.2}$\\

      \cmidrule(l){2-8}
      \multirow{2}{*}{\mpqa}
      & XLM-R dependency & 49.3 & \hspace{-0.5em}*56.9 & 47.6 & 30.5 & 18.9  & \\
      & Frozen PERIN & 44.0 & 49.0 & 46.6 & 30.7 & \!23.1 & $^{\pm1.0}$ \\

      \cmidrule(l){2-8}
      \multirow{2}{*}{\dsu}
    & XLM-R dependency & 26.8 & 33.6 & 36.4 & 22.9 & 18.0  & \\
    & Frozen PERIN & 13.8 & 37.3 & 33.2 & 24.5 & \!21.3 & $^{\pm2.9}$ \\

    \bottomrule
    \end{tabular}
    }
    \caption{Results from comparable experiments, where the dependency graph model (XLM-R dependency) and frozen PERIN models use the same input and similar number of trainable parameters.
    *~marks significant difference, determined by bootstrap (see Appendix~\ref{sec:appendix:bootstrap}).
    }
    \label{tab:ablation_results}
\end{table}

\section{Analysis}


There are a number of architectural differences between the dependency parsing approaches compared above.
In this section, we aim to isolate the effect of predicting intermediate dependency graphs vs.\ directly predicting sentiment graphs by creating more comparable dependency\footnote{We do not use the model from \newcite{peng2021sparse} as the code is not available.} and PERIN models.
We adapt the dependency model from \newcite{barnes-etal-2021-structured} by removing the token, lemma, and POS embeddings and replacing mBERT \cite{devlin-etal-2019-bert} with XLM-R \cite{conneau-etal-2020-unsupervised}.
The `XLM-R dependency' model thus has character LSTM embeddings and token-level XLM-R features.
Since these are not updated during training, for the opinion-tuple `Frozen PERIN' model, we fix the XLM-R weights to make it comparable.

As shown in Table \ref{tab:ablation_results}, predicting the sentiment graph directly leads to an average gain of 3.7 pp. on the Sentiment Graph \F metric.
For extracting the spans of holder, target, and polar expressions, the benefit is less clear.
Here, the PERIN model only outperforms the XLM-R dependency model 5 of 15 times, which seems to confirm that its benefit is at the graph level.
This is further supported by the fact that the highest gains are found on the datasets with the most nested sentiment expressions and dependency arcs lost due to overlap, which are difficult 
to encode in bi-lexical graphs.

\section{Conclusion}

Previous work cast the task of structured sentiment analysis (SSA) as dependency parsing, converting the sentiment graphs into lossy dependency graphs.
In contrast, we here present a novel sentiment parser which predicts sentiment graphs directly from text without reliance on lossy dependency representations.
We adapted a state-of-the-art meaning representation parser and proposed three candidate graph encodings of the sentiment structures. 
Our experimental results suggest that our approach has clear performance benefits, advancing the state of the art on four out of five standard SSA benchmarks.
Specifically, the most direct opinion-tuple encoding provides the highest performance gains.
More detailed analysis of the results shows that the benefits stem from better extraction of global structures, rather than local span prediction. 
Finally, we believe that various structured prediction problems in NLP can similarly be approached in a uniform manner as parsing into directed 
graphs.



\clearpage

\bibliography{anthology,custom}
\bibliographystyle{acl_natbib}

\appendix

\renewcommand{\arraystretch}{1.4}

\begin{table*}[!t]
    \newcommand{\impr}[1]{\underline{\textbf{#1}}}
    \centering
    \catcode`! = 13\def\!{\bfseries}
    \small
    \resizebox{\textwidth}{!}{
    \begin{tabular}{@{}llr@{}lr@{}lr@{}lr@{}lr@{~}r@{}lrr@{}}
    \toprule
  \multirow{2}{*}{\textbf{Dataset}} & \multirow{2}{*}{\textbf{Model}} & \multicolumn{6}{c}{\crulefill~~\textbf{Span \F}~~\crulefill}   & \multicolumn{5}{c}{\crulefill~~\textbf{Sent. graph}~~\crulefill} & \multirow{2}{*}{\textbf{Runtime}} & \multirow{2}{*}{\textbf{\# Params}} \\

    & & 
    \multicolumn{2}{l}{\text{Holder}} & \multicolumn{2}{l}{\text{Target}} & \multicolumn{2}{l}{\text{Exp.}} & \multicolumn{3}{l}{\text{NS\F $\uparrow$}} & \multicolumn{2}{@{}l}{\text{S\F $\uparrow$}} \\
    \midrule
      
      \multirow{4}{*}{\norec}
    & PERIN -- node-centric & 54.9 & $^{\pm4.3}$ & 52.7 & $^{\pm2.0}$ & 57.4 & $^{\pm1.5}$ & 44.8 & $^{\pm1.8}$ & $_{r:\:36.4}^{p:\:46.4}$ & 40.8 & $^{\pm1.5}$ & 9:52 h & 108.9 M\\
    & PERIN -- labeled edge & 59.4 & $^{\pm2.8}$ & 52.0 & $^{\pm2.3}$ & 57.5 & $^{\pm2.7}$ & 44.4 & $^{\pm1.7}$ & $_{r:\:37.7}^{p:\:45.7}$ & 41.1 & $^{\pm1.5}$ & 9:58 h & 109.5 M\\
    & PERIN -- opinion-tuple & 59.2 & $^{\pm1.3}$ & 59.6 & $^{\pm1.3}$ & 61.5 & $^{\pm1.0}$ & 49.4 & $^{\pm1.0}$ & $_{r:\:45.5}^{p:\:42.5}$ & 43.9 & $^{\pm0.9}$ & 9:25 h & 108.1 M\\
    & Frozen PERIN -- opinion-tuple & 50.1 & $^{\pm2.5}$ & 53.8 & $^{\pm1.6}$ & 59.4 & $^{\pm1.0}$ & 44.0 & $^{\pm0.6}$ & $_{r:\:42.2}^{p:\:33.6}$ & 37.4 & $^{\pm0.9}$ & 0:25 h & 23.1 M\\


    \cmidrule(l){2-15}
      \multirow{4}{*}{\basque}
      & PERIN -- node-centric & 57.1 & $^{\pm3.1}$ & 68.7 & $^{\pm1.5}$ & 69.9 & $^{\pm1.0}$ & 61.1 & $^{\pm1.1}$ & $_{r:\:56.8}^{p:\:62.8}$ & 59.7 & $^{\pm1.3}$ & 1:02 h & 87.6 M\\
      & PERIN -- labeled edge & 51.2 & $^{\pm4.7}$ & 66.1 & $^{\pm2.1}$ & 66.0 & $^{\pm1.0}$ & 59.4 & $^{\pm1.2}$ & $_{r:\:55.1}^{p:\:60.1}$ & 57.4 & $^{\pm1.2}$ & 0:57 h & 88.2 M\\
      & PERIN -- opinion-tuple & 57.3 & $^{\pm3.0}$ & 65.1 & $^{\pm2.3}$ & 68.6 & $^{\pm0.3}$ & 59.9 & $^{\pm1.0}$ & $_{r:\:54.7}^{p:\:64.5}$ & 59.2 & $^{\pm0.6}$ & 1:04 h & 86.9 M \\ 
      & Frozen PERIN -- opinion-tuple & 57.0 & $^{\pm10.4}$ & 61.1 & $^{\pm3.2}$ & 65.1 & $^{\pm3.9}$ & 55.5 & $^{\pm2.9}$ & $_{r:\:48.8}^{p:\:56.3}$ & 52.2 & $^{\pm3.2}$ & 0:06 h & 0.7 M\\ 

     \cmidrule(l){2-15}
      \multirow{4}{*}{\cat}
      & PERIN -- node-centric & 57.1 & $^{\pm2.0}$ & 73.8 & $^{\pm2.5}$ & 74.2 & $^{\pm1.6}$ & 68.4 & $^{\pm2.6}$ & $_{r:\:62.9}^{p:\:69.9}$ & 66.2 & $^{\pm2.1}$ & 1:17 h& 87.6 M\\
      & PERIN -- labeled edge & 48.9 & $^{\pm4.3}$ & 72.1 & $^{\pm0.9}$ & 72.6 & $^{\pm1.1}$ & 67.1 & $^{\pm1.6}$ & $_{r:\:61.8}^{p:\:69.5}$ & 65.4 & $^{\pm1.6}$ & 1:13 h & 88.2 M\\
      & PERIN -- opinion-tuple & 46.1 & $^{\pm3.0}$ & 74.4 & $^{\pm1.0}$ & 72.9 & $^{\pm0.5}$ & 68.4 & $^{\pm1.5}$ & $_{r:\:61.6}^{p:\:73.6}$ & 67.0 & $^{\pm1.2}$ & 1:20 h & 86.9 M\\
      & Frozen PERIN -- opinion-tuple & 48.1 & $^{\pm6.4}$ & 65.5 & $^{\pm1.8}$ & 69.2 & $^{\pm5.5}$ & 62.2 & $^{\pm2.7}$ & $_{r:\:56.0}^{p:\:64.7}$ & 59.9 & $^{\pm2.5}$ & 0:07 h & 0.7 M\\ 


      \cmidrule(l){2-15}
      \multirow{4}{*}{\mpqa}
      & PERIN -- node-centric & 58.2 & $^{\pm1.3}$ & 60.8 & $^{\pm0.9}$ & 56.8 & $^{\pm1.1}$ & 35.3 & $^{\pm1.3}$ & $_{r:\:28.7}^{p:\:34.5}$ & 31.4 & $^{\pm1.4}$ & 6:46 h & 107.7 M\\
      & PERIN -- labeled edge & 57.1 & $^{\pm2.0}$ & 54.8 & $^{\pm1.6}$ & 55.2 & $^{\pm1.1}$ & 33.1 & $^{\pm0.4}$ & $_{r:\:26.4}^{p:\:35.7}$ & 30.3 & $^{\pm0.5}$ & 7:16 h & 109.6 M\\
      & PERIN -- opinion-tuple & 56.0 & $^{\pm0.6}$ & 64.2 & $^{\pm1.7}$ & 51.7 & $^{\pm2.8}$ & 42.1 & $^{\pm0.8}$ & $_{r:\:30.1}^{p:\:44.3}$ & 35.8 & $^{\pm0.6}$ & 6:43 h & 108.1 M\\
      & Frozen PERIN -- opinion-tuple & 42.0 & $^{\pm3.8}$ & 48.1 & $^{\pm1.7}$ & 46.6 & $^{\pm2.6}$ & 28.1 & $^{\pm2.2}$ & $_{r:\:20.8}^{p:\:24.3}$ & 22.2 & $^{\pm1.5}$ & 0:37 h & 23.1 M\\ 


      \cmidrule(l){2-15}
      \multirow{4}{*}{\dsu}
    & PERIN -- node-centric & 0.0 & $^{\pm0.0}$ & 41.5 & $^{\pm4.3}$ & 40.3 & $^{\pm2.6}$ & 27.2 & $^{\pm2.0}$ & $_{r:\:16.9}^{p:\:33.4}$ & 22.4 & $^{\pm1.3}$ & 2:31 h & 107.7 M\\
    & PERIN -- labeled edge & 0.0 & $^{\pm0.0}$ & 46.5 & $^{\pm1.8}$ & 41.9 & $^{\pm3.4}$ & 28.4 & $^{\pm2.7}$ & $_{r:\:17.8}^{p:\:33.2}$ & 23.1 & $^{\pm2.0}$ & 2:37 h & 109.6 M\\
    & PERIN -- opinion-tuple & 12.0 & $^{\pm11.0}$ & 50.9 & $^{\pm4.7}$ & 42.6 & $^{\pm3.9}$ & 34.9 & $^{\pm4.1}$ & $_{r:\:22.6}^{p:\:39.5}$ & 28.6 & $^{\pm3.5}$ & 2:30 h & 108.1 M\\
    & Frozen PERIN -- opinion-tuple & 0.0 & $^{\pm0.0}$ & 42.7 & $^{\pm4.8}$ & 35.9 & $^{\pm3.3}$ & 26.0 & $^{\pm3.3}$ & $_{r:\:16.3}^{p:\:29.1}$ & 20.3 & $^{\pm2.0}$ & 0:22 h & 23.1 M\\ 

    \bottomrule
    \end{tabular}
    }
    \caption{Development scores of all our models from the main section of this paper. S\F scores are extended by the average precision and recall values. We also show the runtime of a single model and the number of trainable parameters.
    }
    \label{tab:dev-results}
\end{table*}

\renewcommand{\arraystretch}{1.0}

\begin{table}[!t]
    \newcommand{\impr}[1]{\underline{\textbf{#1}}}
    \centering
    \catcode`! = 13\def\!{\bfseries}
    \resizebox{\columnwidth}{!}{
    \begin{tabular}{@{}llrrrrr@{}}
    \toprule
    \multirow{2}{*}{\textbf{Dataset}} && \multicolumn{3}{c}{\crulefill~~\textbf{Span \F}~~\crulefill}   & \multicolumn{2}{c@{}}{\crulefill~~\textbf{Sent. graph}~~\crulefill} \\
        & & H. & T. & E. &  NS\F & S\F \\
    \midrule

      \multirow{2}{*}{\mpqa}
      & original & 44.7 & 51.3 & 45.7 & 25.4 & 15.0 \\
      & new data & 49.3 & 56.9 & 47.6 & 30.5 & 18.9 \\
      & $\Delta$ & $+$4.6 & $+$5.6 & $+$1.9 & $+$5.1 & $+$4.9\\

      \cmidrule(l){2-7}
      \multirow{2}{*}{\dsu}
    & original & 21.0 &  22.6 & 35.2 & 24.0 & 21.0 \\
    & new data & 26.8 &  33.6 & 36.4 & 22.9 & 18.0 \\
    & $\Delta$ & $+$5.8 & $+$11.0 & $+$1.3 & $-$1.1 & $-$3.0 \\

    \bottomrule
    \end{tabular}
    }
    \caption{Results comparing the XLM-R dependency model on the original \mpqa and \dsu data, and the new data.}
    \label{tab:new_data}
\end{table}

\section{Changes to datasets}

We found out that the official data published at {\small\url{https://competitions.codalab.org/competitions/33556}} was slightly changed from the data used in previous related work. Specifically the \mpqa and \dsu datasets had removed a number of errors resulting from the annotation and from the conversion scripts used to create the sentiment graph representations. We re-run the experiments for the comparable baseline model and show the performance differences in Table \ref{tab:new_data}.

\section{Bootstrap Significance Testing}
\label{sec:appendix:bootstrap}

In order to see whether the performance differences for the experiments are  significant, we do bootstrap significance testing \citet{berg-kirkpatrick2012empirical}, combining two variations.
First, we resample the test sets with replacement from all 5 runs together, $b=1\,000\,000$ times, setting the threshold at ${p=0.05}$.
Additionally, we test each pair out of the $5 \times 5$ combinations for all runs, resampling the test set with replacement $b=100\,000$ times, setting the threshold again at $p=0.5$.
When one system is significantly better in 15 out of the 25 comparisons, and additionally significantly better in the first joint test, we finally mark it as significantly better.

\section{Results on development data}
\label{sec:dev-results}

To make any future comparison of our approach easier, we show the development scores of all reported models in Table \ref{tab:dev-results}.

\section{Training details}
\label{sec:training-details}

Generally, we follow the training regime described in the original PERIN paper \cite{samuel-straka-2020-ufal}. The trainable parameters are updated with the AdamW optimizer \citep{loshchilov2017decoupled}, and their learning rate is linearly warmed-up for the first 10\% of the training to improve stability, and then decayed with a cosine schedule. The XLM-R parameters are updated with a lower learning rate and higher weight decay to improve generalization. Similarly to PERIN, we freeze the embedding parameters for increased efficiency and regularization. Following the finding by \newcite{zhang2021revisiting}, we use small learning rates and fine-tune for a rather long time to increase the training stability. Unlike the authors of PERIN, we did not find any benefits from a dynamic scaling of loss weights \cite{chen2018gradnorm}, so we simply set all loss weights to constant $1.0$.

We trained our models on a single Nvidia P100 with 16GB memory, the runtimes are given in Table \ref{tab:dev-results}. We made five runs from different seeds for each reported value to better estimate the expected error. The hyperparameter configurations for all runs follow, please consult the released code for more details and context:  {\small\url{github.com/jerbarnes/direct_parsing_to_sent_graph}}.

\paragraph{General hyperparameters}
\hspace{0.25em}

\begin{lstlisting}[language=Python, basicstyle=\ttfamily\footnotesize]
batch_size = 16                    
beta_2 = 0.98                      
char_embedding = True              
char_embedding_size = 128          
decoder_learning_rate = 6.0e-4       
decoder_weight_decay = 1.2e-6      
dropout_anchor = 0.4              
dropout_edge_label = 0.5           
dropout_edge_presence = 0.5        
dropout_label = 0.85                
dropout_transformer = 0.25          
dropout_transformer_attention = 0.1
dropout_word = 0.1                 
encoder = "xlm-roberta-base"       
encoder_freeze_embedding = True    
encoder_learning_rate = 6.0e-6      
encoder_weight_decay = 0.1       
epochs = 200                       
focal = True                       
freeze_bert = False                
hidden_size_ff = 4 * 768           
hidden_size_anchor = 256           
hidden_size_edge_label = 256       
hidden_size_edge_presence = 256    
layerwise_lr_decay = 0.9           
n_attention_heads = 8              
n_layers = 3                       
query_length = 1                   
pre_norm = True                             
\end{lstlisting}

\vspace{0.25em}
\paragraph{\norec node-centric hyperparameters}
\hspace{0.25em}

\begin{lstlisting}[language=Python, basicstyle=\ttfamily\footnotesize]
graph_mode = "node-centric"
query_length = 2
\end{lstlisting}

\vspace{0.25em}
\paragraph{\norec labeled-edge hyperparameters}
\hspace{0.25em}  \begin{lstlisting}[language=Python, basicstyle=\ttfamily\footnotesize]
graph_mode = "labeled-edge"
query_length = 2
\end{lstlisting}

\vspace{0.25em}
\paragraph{\norec opinion-tuple hyperparameters}
\hspace{0.25em}  \begin{lstlisting}[language=Python, basicstyle=\ttfamily\footnotesize]
graph_mode = "opinion-tuple"
\end{lstlisting}

\vspace{0.25em}
\paragraph{\norec frozen opinion-tuple hyperparameters}
\hspace{0.25em}  \begin{lstlisting}[language=Python, basicstyle=\ttfamily\footnotesize]
graph_mode = "opinion-tuple"
freeze_bert = True
batch_size = 8
decoder_learning_rate = 1.0e-4
dropout_transformer = 0.5
epochs = 50
\end{lstlisting}

\vspace{0.25em}
\paragraph{\basque node-centric hyperparameters}
\hspace{0.25em}  \begin{lstlisting}[language=Python, basicstyle=\ttfamily\footnotesize]
graph_mode = "node-centric"
query_length = 2
n_layers = 0
\end{lstlisting}

\vspace{0.25em}
\paragraph{\basque labeled-edge hyperparameters}
\hspace{0.25em}  \begin{lstlisting}[language=Python, basicstyle=\ttfamily\footnotesize]
graph_mode = "labeled-edge"
query_length = 2
n_layers = 0
\end{lstlisting}

\vspace{0.25em}
\paragraph{\basque opinion-tuple hyperparameters}
\hspace{0.25em}  \begin{lstlisting}[language=Python, basicstyle=\ttfamily\footnotesize]
graph_mode = "opinion-tuple"
n_layers = 0
\end{lstlisting}

\vspace{0.25em}
\paragraph{\basque frozen opinion-tuple hyperparameters}
\hspace{0.25em}  \begin{lstlisting}[language=Python, basicstyle=\ttfamily\footnotesize]
graph_mode = "opinion-tuple"
freeze_bert = True
n_layers = 0
epochs = 50
\end{lstlisting}

\vspace{0.25em}
\paragraph{\cat node-centric hyperparameters}
\hspace{0.25em}  \begin{lstlisting}[language=Python, basicstyle=\ttfamily\footnotesize]
graph_mode = "node-centric"
query_length = 2
n_layers = 0
\end{lstlisting}

\vspace{0.25em}
\paragraph{\cat labeled-edge hyperparameters}
\hspace{0.25em}  \begin{lstlisting}[language=Python, basicstyle=\ttfamily\footnotesize]
graph_mode = "labeled-edge"
query_length = 2
n_layers = 0
\end{lstlisting}

\vspace{0.25em}
\paragraph{\cat opinion-tuple hyperparameters}
\hspace{0.25em}  \begin{lstlisting}[language=Python, basicstyle=\ttfamily\footnotesize]
graph_mode = "opinion-tuple"
n_layers = 0
\end{lstlisting}

\vspace{0.25em}
\paragraph{\cat frozen opinion-tuple hyperparameters}
\hspace{0.25em}  \begin{lstlisting}[language=Python, basicstyle=\ttfamily\footnotesize]
graph_mode = "opinion-tuple"
freeze_bert = True
n_layers = 0
epochs = 50
\end{lstlisting}

\vspace{0.25em}
\paragraph{\mpqa node-centric hyperparameters}
\hspace{0.25em}  \begin{lstlisting}[language=Python, basicstyle=\ttfamily\footnotesize]
graph_mode = "node-centric"
decoder_learning_rate = 1.0e-4
query_length = 2
\end{lstlisting}

\vspace{0.25em}
\paragraph{\mpqa labeled-edge hyperparameters}
\hspace{0.25em}  \begin{lstlisting}[language=Python, basicstyle=\ttfamily\footnotesize]
graph_mode = "labeled-edge"
decoder_learning_rate = 1.0e-4
query_length = 2
\end{lstlisting}

\vspace{0.25em}
\paragraph{\mpqa opinion-tuple hyperparameters}
\hspace{0.25em}  \begin{lstlisting}[language=Python, basicstyle=\ttfamily\footnotesize]
graph_mode = "opinion-tuple"
\end{lstlisting}

\vspace{0.25em}
\paragraph{\mpqa frozen opinion-tuple hyperparameters}
\hspace{0.25em}  \begin{lstlisting}[language=Python, basicstyle=\ttfamily\footnotesize]
graph_mode = "opinion-tuple"
freeze_bert = True
batch_size = 8
decoder_learning_rate = 1.0e-4
dropout_transformer = 0.5
epochs = 50
\end{lstlisting}

\vspace{0.25em}
\paragraph{\dsu node-centric hyperparameters}
\hspace{0.25em}  \begin{lstlisting}[language=Python, basicstyle=\ttfamily\footnotesize]
graph_mode = "node-centric"
decoder_learning_rate = 1.0e-4
query_length = 2
\end{lstlisting}

\vspace{0.25em}
\paragraph{\dsu labeled-edge hyperparameters}
\hspace{0.25em}  \begin{lstlisting}[language=Python, basicstyle=\ttfamily\footnotesize]
graph_mode = "labeled-edge"
decoder_learning_rate = 1.0e-4
query_length = 2
\end{lstlisting}

\vspace{0.25em}
\paragraph{\dsu opinion-tuple hyperparameters}
\hspace{0.25em}  \begin{lstlisting}[language=Python, basicstyle=\ttfamily\footnotesize]
graph_mode = "opinion-tuple"
\end{lstlisting}

\vspace{0.25em}
\paragraph{\dsu frozen opinion-tuple hyperparameters}
\hspace{0.25em}  \begin{lstlisting}[language=Python, basicstyle=\ttfamily\footnotesize]
graph_mode = "opinion-tuple"
freeze_bert = True
batch_size = 8
decoder_learning_rate = 1.0e-4
dropout_transformer = 0.5
epochs = 50
\end{lstlisting}


\end{document}